\documentclass[sigconf]{acmart-preprint}

\AtBeginDocument{%
  \providecommand\BibTeX{{%
    \normalfont B\kern-0.5em{\scshape i\kern-0.25em b}\kern-0.8em\TeX}}}

\copyrightyear{2023} 
\acmYear{2023} 
\setcopyright{acmlicensed}\acmConference[SIGIR '23]{Proceedings of the 46th International ACM SIGIR Conference on Research and Development in Information Retrieval}{July 23--27, 2023}{Taipei, Taiwan}
\acmBooktitle{Proceedings of the 46th International ACM SIGIR Conference on Research and Development in Information Retrieval (SIGIR '23), July 23--27, 2023, Taipei, Taiwan}
\acmPrice{15.00}
\acmDOI{10.1145/3539618.3591861}
\acmISBN{978-1-4503-9408-6/23/07}

\begin{document}

\title{Contextual Multilingual Spellchecker for User Queries}


\author{Sanat Sharma}
\email{sanatsha@adobe.com}
\affiliation{%
  \institution{Adobe Inc.}
  \country{USA}
}

\author{Josep Valls-Vargas}
\email{jvallsvargas@adobe.com}
\affiliation{%
  \institution{Adobe Inc.}
  \country{USA}
}

\author{Tracy Holloway King}
\email{tking@adobe.com}
\affiliation{%
  \institution{Adobe Inc.}
  \country{USA}
}

\author{Francois Guerin}
\email{guerin@adobe.com}
\affiliation{%
  \institution{Adobe Inc.}
  \country{USA}
}

\author{Chirag Arora}
\email{charora@adobe.com}
\affiliation{%
  \institution{Adobe Inc.}
  \country{USA}
}


\begin{abstract}
Spellchecking is one of the most fundamental and widely used search features. Correcting  incorrectly spelled user queries  not only enhances the user experience but  is  expected by the user. However, most widely available spellchecking solutions are either  lower accuracy than state-of-the-art solutions or too slow to be used for search use cases where latency is a key requirement. Furthermore, most innovative recent architectures focus on English  and are not trained in a multilingual fashion and are trained for spell correction in longer text, which is a different paradigm from spell correction for user queries, where  context is sparse (most  queries are 1--2 words long). Finally, since most enterprises have unique vocabularies such as product names, off-the-shelf spelling solutions fall short of  users’ needs.  

In this work, we build a multilingual spellchecker that is extremely fast and scalable and that  adapts its vocabulary and hence speller output based on a specific product’s needs. Furthermore  our speller out-performs general purpose spellers by a wide margin on in-domain datasets. Our multilingual speller is used in search in Adobe products, powering autocomplete in various applications.
 
\end{abstract}

\ccsdesc[500]{Applied computing~Document management and text processing}

\keywords{spellcheck, spell correction, neural networks, query processing}


\maketitle

\section{Introduction}

Spell correction is a widely studied problem in search and NLP research. Spellcheckers generally comprise two parts: creating a list of candidate corrections and ranking those candidates. Most widely used spellcheckers are built for  English  and utilize behavioral \cite{langinde} and/or contextual signals \cite{choudhury-etal-2007-difficult} for ranking the suggested corrections. Recent works have also utilized other extrinsic data such as search results \cite{chen-etal-2007-improving} or public domain multi-word datasets \cite{gao-etal-2010-large} as ranking signals. Although most spellers are built for English, some works have developed custom spellers for non-English languages such as Bengali \cite{bengalistemmer} or Dutch \cite{doubledutch}. These works are hard to scale across multiple languages since they are language specific.
Most of the work for spell correction has been around correction in sentences or paragraphs where context is plentiful. In such cases, neural models such as transformers and LSTMs  perform well since they capture textual context  \cite{denoisingtransformer}. However, these systems are usually slower than their frequentists counterparts and do not show much improvement in search query cases where textual context is minimal.

Our work  takes a best-of-both-worlds approach: We utilize contextual signals such as  search results, behavioral data, and phonetic signals to suggest candidates, while incorporating a small neural model for ranking. In addition, we use a suggestion model that is language agnostic and can scale to multiple languages. 

We divide the speller into four components: a behavioral data analysis pipeline to finetune the downstream components; a product specific rule engine to correct common errors and provide editorial overrides; a suggester that takes in user queries and suggests potential replacements for incorrectly spelled tokens; and a neural ranker that calculates the probability of the suggested tokens. We evaluate our speller on both general purpose and product specific domains and showcase significant improvement over current methods. 

Our approach is currently  used in production by the autocomplete feature in Adobe search and is being integrated in  Adobe Express and Adobe Stock for online spell correction. 

Our main contributions and business impact are: 
\begin{enumerate}
    \item A novel approach for creating a fast, multilingual spellchecker for search queries 
    \item A novel, low latency architecture for deploying and scaling the spellchecker 
    \item Significant improvement over widely available state-of-the-art spellcheckers for short user queries 
\end{enumerate}

\section{Training Datasets}
\label{sec:data}

Finding public spellcheck datasets is surprisingly hard, with very few benchmarks  available for validation. Furthermore, since we require data for training our models, we decided to employ a bootstrapping approach for dataset generation and leveraged crowd workers for manual curation. This section describes how we created the training data, as well as some datasets used for initial internal evaluation. The evaluation datasets are described in section \ref{sec:eval}.

\subsection{Artificially Generated Query Dataset }

We extracted user queries from search over Adobe Stock images for English, French and German locales for analysis. Since we use full queries, the model has some context for multi-word queries. 

{\bf Data Preprocessing}: We removed  queries with spelling errors from the dataset by applying the updated Hunspell\footnote{http://hunspell.github.io/} dictionaries to check for spelling errors and then had the remaining queries reviewed by crowd workers. This created our ground truth dataset. 

{\bf Artificial Injection of Errors}: Most spelling errors are due to one of the following reasons: missing a letter, adding a letter, typing an incorrect letter. To create our artificial dataset, for each query in the list of correctly spelled queries, we injected one or more spelling errors using one of the following techniques in a probability-weighted fashion:

\begin{enumerate}
\item Change the order of letters  (e.g.\ "change" to "chnage"; "check" to "chekc"). This the most common spelling error. 

\item Remove or add a vowel (e.g.\  "malleable" to "mallable" or "malleiable"). 

\item Add an additional character  (e.g.\  "fresh" to "freshh" or "frersh"). 

\item Replace a character with another character  (e.g.\  "fresh" to "frash" or "frwsh").

\item Replace accented characters and their unaccented counterparts with another character in the same class (e.g.\  "français" to "francais"; "wörter" to "worter"). 

\item For words with two identical letters in a row, have only one letter  (e.g.\ change "happiness" to "hapiness"). 
\end{enumerate}

\noindent The artificial errors were patterned on real life errors and were weighted at a ratio of 7:5:4:2:7:2 respectively. For addition of vowels, only vowels that usually follow one another were chosen, e.g.\ for   `e',   `i' was much more likely to be added  than `u'. Each query in the example set had one or more errors injected into them.

Our final artificial dataset size is shown in in Table \ref{tab:trainingsize}.

\begin{table}[htb]
  \begin{tabular}[t]{lll}
English Queries & $\sim$1.5M\\
French Queries &   $\sim$1.2M \\
German Queries & $\sim$1.4M\\
\end{tabular}
\caption{Training data size}
\label{tab:trainingsize}
\end{table}

Table \ref{tab:examplequeries} shows example input queries and their artificially misspelled training counterparts. Only some words in the query have added errors so that the model also learns to recognize correctly spelled words.

\begin{table}[htb]
\begin{tabular}[t]{p{2cm}p{2cm}p{3cm}}  
Query & Error Tokenized Query &Error Type \\\hline\hline
atlantic  & [agtlantic,  & {\sc  letter\_add\_remove} \\
 mackerel &  mackrel] &{\sc  vowel\_add\_remove} \\[1ex]
 burgundy & [burgundy, & {\sc double\_add\_remove} \\
 background &  backgrround] & \\[1ex]
glacier national & [glaicer, & {\sc letter\_order} \\
park and hike  & natoinal, 0ark, & {\sc letter\_order}\\
&and, hik] & {\sc letter\_change}\\&& {\sc letter\_add\_remove} \\[1ex]
medal icon & [,edal, icon] & {\sc letter\_change} \\
\end{tabular}
\caption{Examples of spelling errors introduced to naturally occurring, correctly spelled Adobe Stock queries. Note that punctuation marks and numbers can substitute for letters.}
\label{tab:examplequeries}
\end{table}

\subsection{Birkbeck Corpus }

The Birkbeck corpus\footnote{http://www.dcs.bbk.ac.uk/$\sim$ROGER/corpora.html}\ contains 36,133 misspellings of 6,136 words. It is an amalgamation of errors taken from the native-speaker section (British and American writers) of the Birkbeck spelling error corpus, a collection of spelling errors gathered from various sources, available with detailed documentation from the Oxford Text Archive.\footnote{http://ota.ahds.ac.uk/      }\ It includes the results of spelling tests and errors from free writing, primarily from schoolchildren, university students and adult literacy students. We utilize 18,295 misspellings from Birkbeck as part of our English training dataset.

\subsection{Commonly Misspelled Word Corpora }

The Aspell \cite{aspell} corpus contains  $\sim$1500 common misspellings.
Wiki\-pe\-dia\footnote{https://en.wikipedia.org/wiki/Commonly\_misspelled\_English\_words}\  lists commonly misspelled words. We used these to mine for queries in our domain that feature these misspelled words and for internal evaluation for model selection.

\section{Model}

\begin{figure*}[htb]
    \centering
    \includegraphics[width=5in]{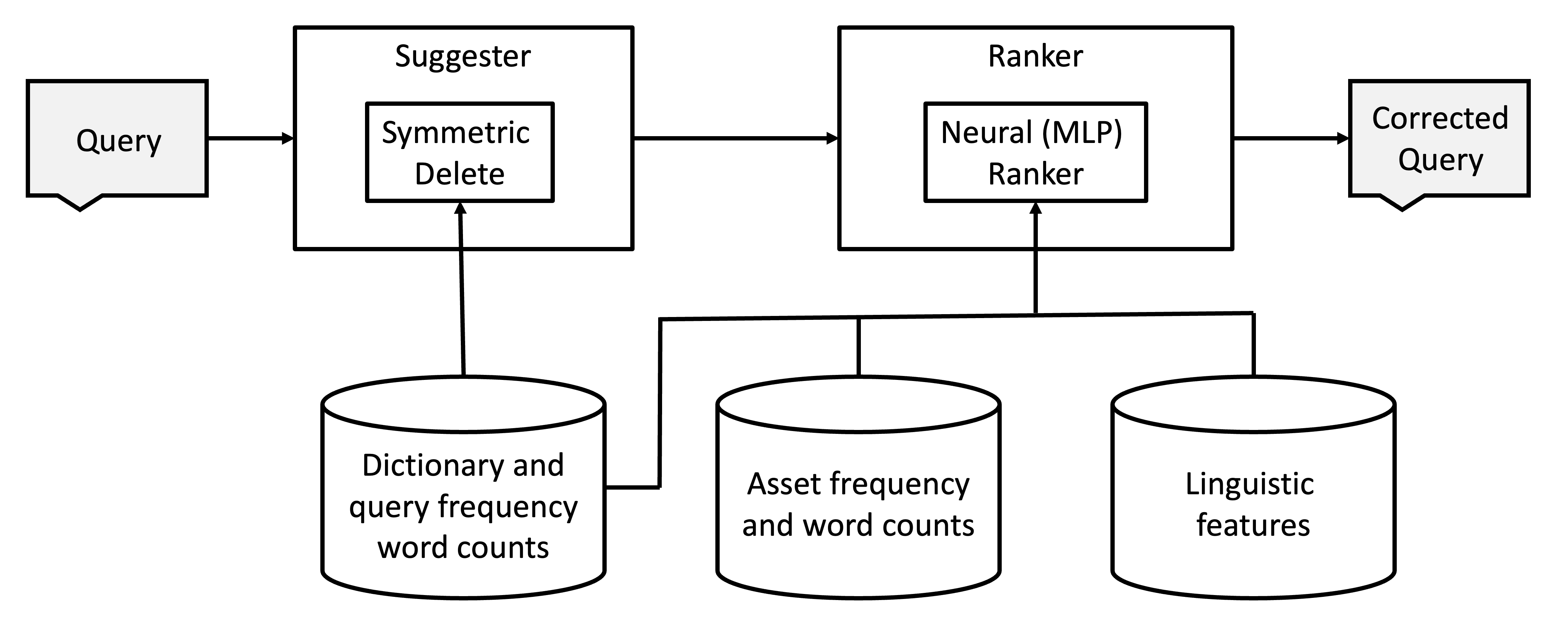}
    \caption{Model architecture of the speller}
    \label{fig:model_architecture}
\end{figure*}

Following common practice, we divide the spellcheck model into two modules: a suggester module and a ranker module. The suggester module takes in the user query and suggests possible correction tokens for any incorrectly spelled tokens. The ranker module ranks the suggestions and outputs the most probable candidate. This is shown in Figure\ \ref{fig:model_architecture}.

\subsection{Symmetric Delete Suggester}

We utilize the Symmetric Delete\footnote{https://github.com/wolfgarbe/SymSpell} \cite{symspell} algorithm for our suggester module. Symmetric Delete generates a  permutation index for words in the dictionary at index time. Instead of calculating transposes + replaces + inserts + deletes at runtime, Symmetric Delete only calculates deletes of the index dictionary. The symmetric delete suggester has two key advantages: 
\begin{itemize}
    \item \textbf{Latency}: The module is extremely fast for up to 2 edit distances, with an average of $\sim$1ms latency. This is critical for query spell correction. The speed comes from inexpensive delete-only edit candidate generation and pre-calculation. 
    \item \textbf{Language Agnostic}: The module is language agnostic, not requiring language specific characteristics to generate suggestions.
\end{itemize}

\paragraph{Index Time Operation}
At index time, we utilize a dictionary of correct words and generate the symmetric delete index from those. The dictionary of correct words is generated from known language dictionaries, including FastText \cite{DBLP:journals/corr/BojanowskiGJM16} word dictionaries, Adobe-specific product terms (e.g.\ product names, file extensions) and behavioral data (e.g.\ popular queries). 
The addition of custom vocabulary is important because most enterprises have custom language that is not  supported by the  open source dictionaries.

\paragraph{Runtime Operation}
At runtime, given a user query, we first check if the query is correctly spelled. If it is incorrect, we find all candidates within 1 edit distance. If <3 candidates are generated, we then utilize 2 edit distance suggestions. This balances  speed and precision, as increasing the edit distance leads to more suggestions but higher latency. In our analysis of Adobe user queries, we found that  88\% of  spelling errors are 1 edit distance away. So, 2 edit distance suggestions are used sparingly.

\subsection{Neural Ranker}

We utilize a neural network to rank the suggestions from the suggester module. Due to our low latency requirements, we use a multilayer perceptron network (MLP) rather than recurrent neural nets or transformers. Our MLP consists of 5 fully connected layers, with dropout and batch normalization. Since MLPs do not do well at token level understanding,  we utilize the features for each suggestion rather than the tokens themselves in order to improve performance on unseen words (i.e.\ unique spelling errors).

The features we utilize for each suggestion are below. All features were scaled and normalized (0--1) before being fed to the neural network. 
\begin{itemize}
    \item {\bf Word Count}: In most cases, we want to recommend  more common words. We store the number of occurrences of each word in the query set. The word counts  vary based on application, enabling per-application suggestions.
    \item {\bf Asset Frequency}: In most cases, we want to correct to a word which  retrieves more search results. For each  word, we store the number of assets associated with it. This feature is application specific.
    \item {\bf Download Count}: Query success is indicated by downloads in Adobe Stock.  We store the number of downloads for the first 100 (first page) results for each word. This feature is only used on  Adobe Stock.
    \item {\bf Levenshtein Distance}: Standard string edit distance measurement \cite{Levenshtein_SPD66}.
    \item {\bf Language Locale}: Language of the locale the query is issued on (e.g.\ French, Japanese).
    \item {\bf Application}: Which application the query is scoped to (e.g.\ Adobe Stock, Adobe Express).
    \item {\bf Phonetic Similarity}: For misspellings where the misspelled word is phonetically correct  (e.g. muzeem vs.\ museum), this feature helps focus on phonetically similar corrections.\footnote{This feature was only utilized for  English.}
\end{itemize}

\section{Service Architecture}

\begin{figure*}
    \centering
    \includegraphics[width=5in]{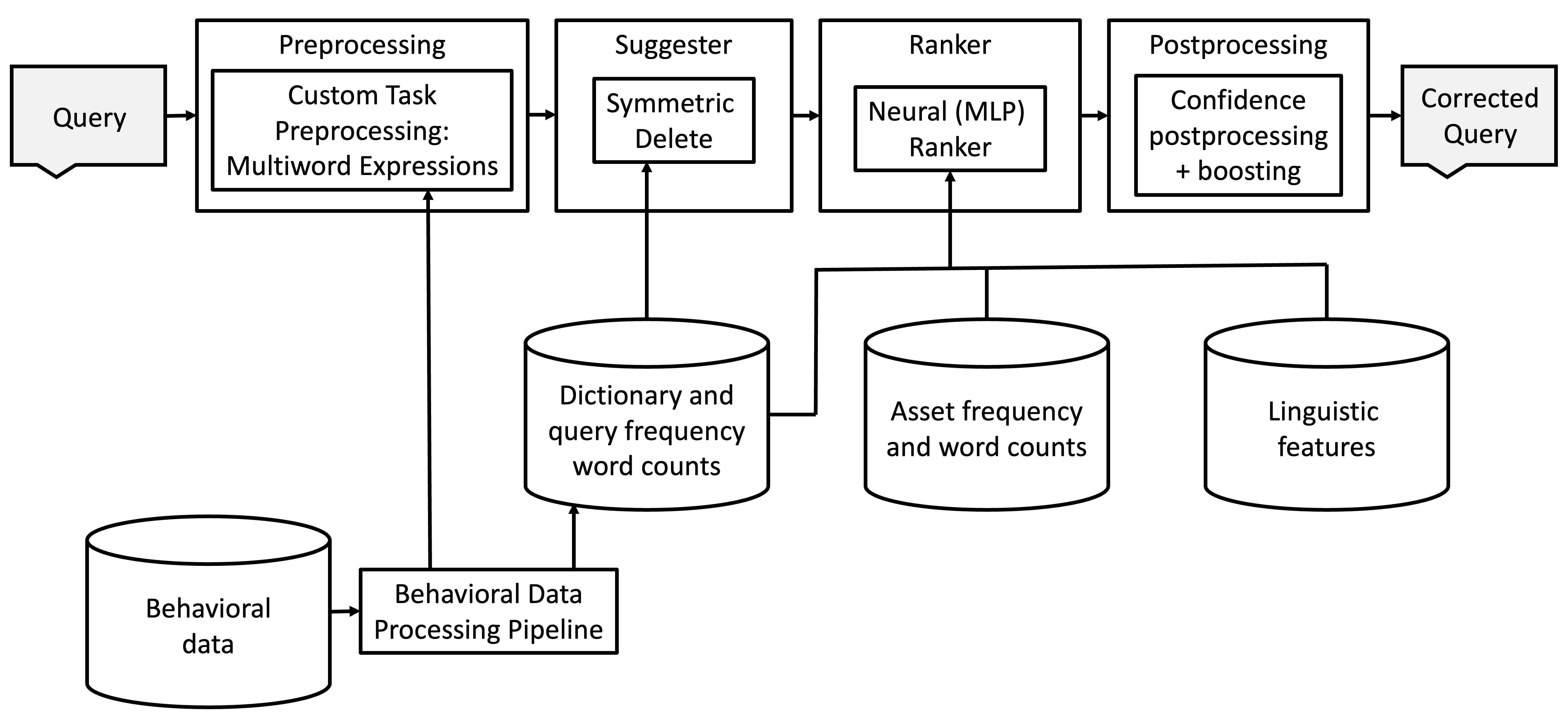}
    \caption{Spellcheck Service Architecture. The MWE module handles  task-specific multi-word expressions before the suggester and ranker are called. Behavioral pipelines keep features  updated. The postprocessor enables task-specific confidence boosting.}
    \label{fig:sys_design}
\end{figure*}

In order to serve, scale and maintain low latency for the spellchecker, we implemented a novel architecture (Figure \ref{fig:sys_design}). The suggester module can struggle with task-specific multi-word errors caused by compounding or decompounding (e.g.\ "creativecloud" (creative cloud) and "photo shop express" (photoshop express)). We created a multi-word expression  (MWE) module that corrects the most custom multi-word errors. This module uses a key-value map based on the most common queries in Adobe products and can be different for different applications. We also created a behavioral pipeline that automatically updates the statistics for the  model features (e.g.\ asset frequency, word count). This  updates the speller  based on user data without the need for extrinsic changes (e.g.\ automatically incorporating   new words like "covid" and "blockchain").

\section{Evaluation}
\label{sec:eval}

We performed several qualitative and quantitative evaluations on a variety of datasets. Here, we present results from two different applications to  demonstrate the ability of the speller to adapt to different query patterns.

\subsection{Adobe Express User Queries}

We performed a quantitative analysis on user queries from  Adobe Express, a web-based product to create assets from templates. We generated a  misspelling dataset from Adobe Express queries by mining queries using the commonly misspelled words from the Wikipedia and Aspell  datasets. Additionally we added synthetic perturbations on the mined queries  based on common misspellings for each of the 3 languages under consideration (English, French, German). Finally, high frequency spelling errors seen in the application were  added via human annotation. There are 6355 queries for English, 1187 for German, and 1128 for  French.

This dataset is very different from the dataset that our model was trained on (section \ref{sec:data}) but uses dictionaries from the same distribution. This gives us  a better representation of real world performance across domains. We tested the performance against     NeuSpell (a state-of-the-art neural spelling model) \cite{DBLP:journals/corr/abs-2010-11085} and Aspell (a widely used speller) \cite{aspell}. As shown in Table \ref{tab:accuracy}, our approach outperforms off-the-shelf state-of-the-art approaches in our specific domain, while taking a fraction of the time (under 1 ms on average as opposed to 40+ ms). 

\begin{table}[htb]
\centering
\begin{tabular}{|l|r|r|r|c|}
\hline
Model & \multicolumn{3}{c|}{Accuracy} & Latency  \\ \cline{2-4}
 & English & French &German & (ms) \\
\hline
Aspell & 51.6\% & 60.8\% & 29.7\% & 40 \\
\hline
Neuspell & 75.5\% & 37.5\% & 36.6\% & 50 \\
\hline
Ours & 81.7\% & 85.0\% & 84.8\% & <1 \\
\hline
\end{tabular}
\caption{Accuracy and latency of different spell correction models on the Adobe Express query dataset}
\label{tab:accuracy}
\end{table}


\subsection{Adobe Creative Cloud Home User Queries}

We performed a qualitative analysis on user queries from Adobe Creative Cloud Home, one of the main gateways for users to search about Adobe products. We utilized English queries from a single day. The evaluation set comprised  7123 unique queries and their frequencies.

We crowd-sourced and manually checked the correctness of the response from the speller. Results are depicted in Table \ref{tab:cchome}.  Nearly 50\% of all unique queries entered by users contained a spelling error,  highlighting the need for a task-specific speller. Most of the common spelling errors revolved around product names with the words "creative" or "acrobat" being spelled incorrectly in many different ways. For this application, having higher boosting for Adobe product name candidates led to better results due to the nature of the queries, highlighting the need for application-specific contextual signals.

\begin{table}[htb]
\centering
\begin{tabular}
{|l|r|r|r|}
\hline
Model & Recall & Precision & Accuracy  \\ 
\hline
Aspell  & 29.5\% & 98.9\% & 45.5\% \\
\hline
Neuspell  & 57.6\% & 84.2\% & 75.7\% \\
\hline
Ours  & 96.4\% & 87.3\% &  82.2\% \\
\hline
\end{tabular}
\caption{Accuracy metrics on the Creative Cloud Home dataset. Recall is the rate of incorrect queries that have been properly corrected. Precision is the rate of corrected queries where the correction is correct. }
\label{tab:cchome}
\end{table}

\section{Conclusions and Next Steps}

In this paper we described a novel approach for creating a fast, multilingual spellchecker for user queries. This includes a novel, low latency architecture for deploying and scaling the spellchecker. The resulting speller shows significant improvement over widely available state-of-the-art spellcheckers for short user queries. 

Next steps focus on two areas. The first is using the English, French, and German spellers to replace the current production query-time spellers given their success in offline spell correction for autocomplete. The second is extending the speller to $\sim$10 and eventually $\sim$35 languages in order to cover the primary languages used in our search applications. This will allow us to use the same high-quality, custom-tuned, low-latency speller for all query spell correction, both offline for autocomplete suggestions and online for user queries.

\section*{Adobe Company Portrait} Adobe Inc.\ enables customers to change the world through digital experiences and creativity. The Adobe search and discovery team supports search and recommendations across customer text, image, video,  and other document types as well as over Adobe Stock assets and Adobe help and tutorials.

\section*{Main Author Bio} 
Sanat Sharma is a senior machine learning engineer at Adobe Inc. He
earned his Master's degree from University of Texas, Austin in 2020, with a focus on NLP. Sanat's work focuses on search improvements and contextual recommendations, and his work has been published at conferences such as CVPR.

\bibliographystyle{ACM-Reference-Format}
\bibliography{SIRIPrefs}

\end{document}